\documentclass[10pt,twocolumn,letterpaper]{article}

\usepackage[pagenumbers]{cvpr} 

\usepackage{booktabs}
\usepackage{makecell}
\usepackage{algorithm}
\usepackage{algpseudocode}
\usepackage{amsmath}
\usepackage{bbding}
\definecolor{cvprblue}{rgb}{0.21,0.49,0.74}
\usepackage[pagebackref,breaklinks,colorlinks,allcolors=cvprblue]{hyperref}

\newcommand{\ysold}[1]{#1}

\def\paperID{606} 
\def\confName{CVPR}
\def\confYear{2026}

\title{FlowPortal: Residual-Corrected Flow for \\Training-Free Video Relighting and Background Replacement\vspace{-0.6em}}

\author{Wenshuo Gao \hspace{12pt} Junyi Fan  \hspace{12pt} Jiangyue Zeng  \hspace{12pt} Shuai Yang$^\textrm{\Envelope}$\\
\normalsize{Wangxuan Institute of Computer Technology, State Key Laboratory of Multimedia Information Processing,} \\\normalsize{Peking University, Beijing, China}\\
{\tt\small \{gaowenshuo, bright\_fjy, zengjy\}@stu.pku.edu.cn \hspace{12pt} williamyang@pku.edu.cn}\vspace{-4mm}
}

\begin{document}
\twocolumn[{%
\vspace{-15mm}
\renewcommand\twocolumn[1][]{#1}%
\maketitle
\centering
\includegraphics[width=0.92\linewidth]{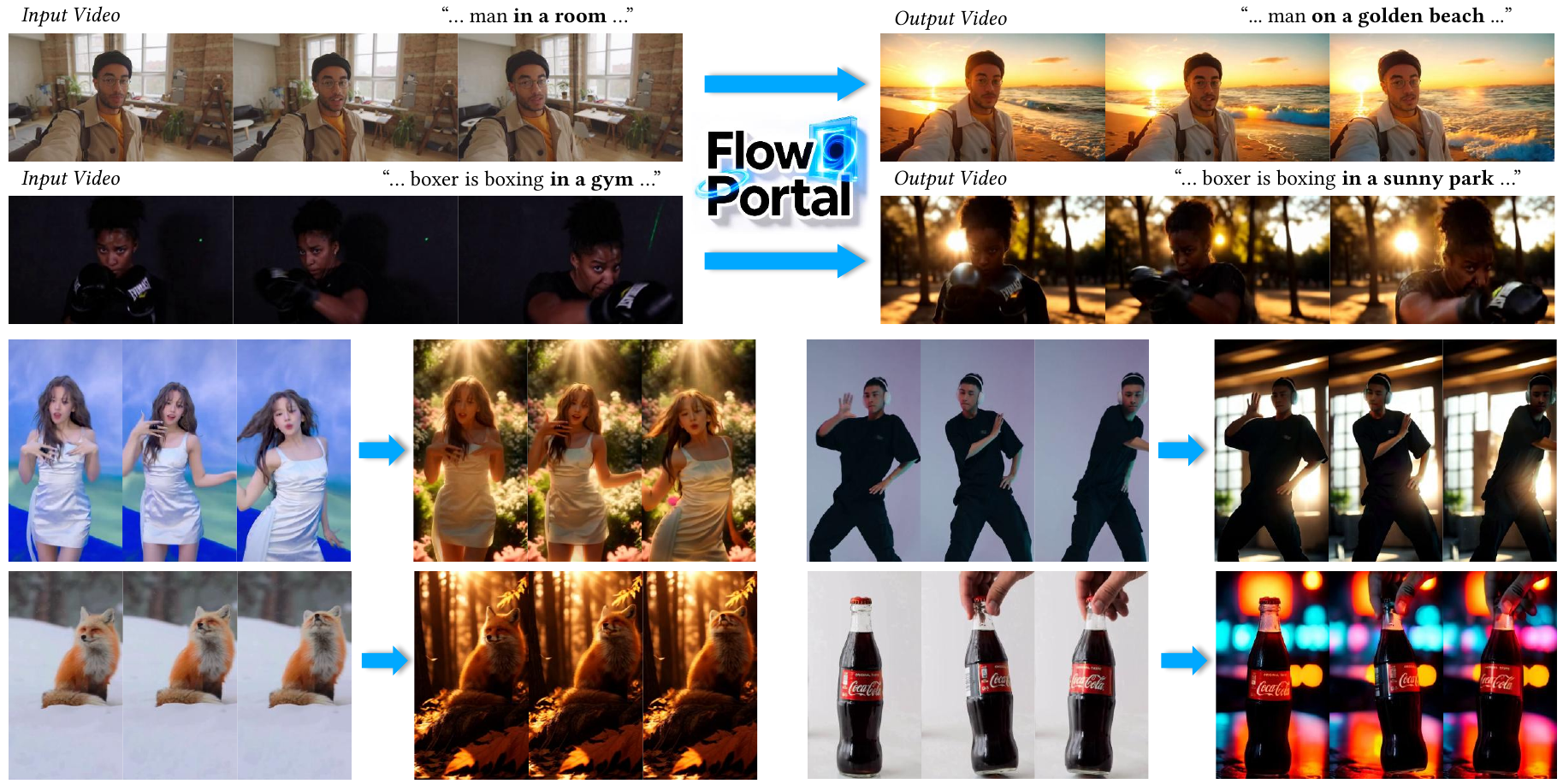}
 \vspace{-2mm}
\captionof{figure}{We propose a novel training-free \textbf{FlowPortal} framework for efficient video background replacement and foreground relighting. 
}
\vspace{3mm}
\label{fig:teaser}
}]
\maketitle

\urlstyle{rm}

\begin{abstract}
Video relighting with background replacement is a challenging task critical for applications in film production and creative media. Existing methods struggle to balance temporal consistency, spatial fidelity, and illumination naturalness. To address these issues, we introduce FlowPortal, a novel training-free flow-based video relighting framework. Our core innovation is a Residual-Corrected Flow mechanism that transforms a standard flow-based model into an editing model, guaranteeing perfect reconstruction when input conditions are identical and enabling faithful relighting when they differ, resulting in high structural consistency. This is further enhanced by a Decoupled Condition Design for precise lighting control and a High-Frequency Transfer mechanism for detail preservation. Additionally, a masking strategy isolates foreground relighting from background pure generation process. Experiments demonstrate that FlowPortal achieves superior performance in temporal coherence, structural preservation, and lighting realism, while maintaining high efficiency. Project Page: \url{https://gaowenshuo.github.io/FlowPortalProject/}.
\end{abstract}    
\section{Introduction}
\label{sec:intro}

Harmonious lighting between the foreground and the background plays a crucial role in determining the realism, visual quality, and aesthetic appeal of a video. The capability to generate a new background that harmonizes with the foreground using adapted lighting, known as the video relighting task with background replacement, has attracted increasing attention due to its broad range of applications in film production, virtual photography, commercial display, and creative media editing. This technology effectively functions as a visual portal, dramatically reducing the need for on-site shooting by enabling flexible scene composition and stylized video generation. Furthermore, this capability holds significant potential for future world-to-world portal applications that seamlessly bridge different visual worlds.


\ysold{Although promising, video relighting with background replacement poses significant challenges due to the need to maintain video quality, temporal consistency, and subject fidelity under new lighting conditions.
Pursuing one goal often compromises others, leading to temporal flickering, loss of detail, unrealistic lighting, or prohibitive computational cost. For example, training-based methods~\cite{lin2025illumicraft,he2025unirelight,fang2025relightvid,zeng2025lumen,mei2025lux,liu2025tc} are resource-intensive and struggle with creating paired dataset and balancing between illumination richness and content fidelity. 
Meanwhile, training-free methods like AnyPortal~\cite{gao2025anyportal} and Light-A-Video~\cite{zhou2025light} that combine pretrained image relighting and video generation models suffer from
temporal inconsistency caused by per-frame relighting as well as misalignment between input and output video due to weak condition control and limited capability for video models.}

We argue that these issues arise from the absence of a cohesive framework that systematically disentangles and controls the core elements of a video: structure, motion, and illumination. To bridge this gap, we introduce a new \textbf{FlowPortal} framework built upon a condition-aware editing model \ysold{through two dedicated mechanisms. 
First, we propose a \textbf{Residual-Corrected Flow} with \textbf{High-Frequency Transfer} for structure and detail preservation.
Second, \textbf{Decoupled Condition Design} is introduced for precise lighting control.} Together, these advancements provide a unified solution that addresses the long-standing challenges of video relighting: temporal continuity, spatial fidelity, illumination naturalness, and operational efficiency.

\ysold{\ysold{Specifically,} our key insight is to exploit the principle of Condition Consistency that every change in the output is driven by a change in the target condition (\ie, illumination).~\ysold{It implies a perfect reconstruction when there is no change in condition.}}
Based on this principle, we introduce a Residual-Corrected Flow mechanism that rephrases a typical flow-based generation model to a novel training-free editing model.
It ensures structural consistency by directing the flow towards perfect reconstruction when the target and source conditions are identical, so that subsequent changes in the conditions like illumination variation can be reflected in the video output.
\ysold{Our Residual-Corrected Flow allows for reusing velocity prediction across timesteps to enhance efficiency, and it processes videos holistically, avoiding per-frame processing to ensure temporal continuity.}


\ysold{To further strengthen the model's fidelity and controllability,} we introduce three key designs: 
First, we propose a Decoupled Condition Design with illumination-specific and agnostic visual-textual condition inputs. This decomposition provides a stable, sufficient, and directional guidance that engineeringly enforces Condition Consistency.
Second, we introduce High-Frequency Transfer, which adheres to Condition Consistency by injecting illumination-agnostic details to solidify fidelity.
Third, we employ masks to isolate and precisely relight the foreground while maintaining a pure background generation process, allowing for a high-quality, contextually natural background replacement.
In summary, our contributions are threefold:
\begin{itemize}
\item We propose a novel training-free FlowPortal framework for coherent and efficient video relighting and background replacement.~\ysold{By introducing Residual-Corrected Flow and Decoupled Condition Design, our framework ensures structural consistency between source and target videos, as well as illumination naturalness.}
\item We introduce a High-Frequency Transfer mechanism within the Residual-Corrected Flow, improving fine detail consistency under new lighting conditions.
\item We propose Masked Residual-Corrected Flow and Masked High-Frequency Transfer to isolate foreground and background processing, enabling high-quality, contextually natural background replacement.
\end{itemize}
\section{Related Works}
\label{sec:related}

\textbf{Image Relighting and Background Replacement.}
Recent advances in generation models have led to successful image relighting methods~\cite{xing2025luminet, pandey2021total, kim2024switchlight, ren2024relightful, zhang2025scaling}. Total Relighting~\cite{pandey2021total} and SwitchLight~\cite{kim2024switchlight} train neural networks to predict surface properties, namely normals and albedo, which are then used to recompute illumination.  Relightful Harmonization~\cite{ren2024relightful} fine-tunes an image diffusion model conditioned on the background to achieve foreground relighting. As the current state-of-the-art method in image relighting, IC-Light~\cite{zhang2025scaling} simultaneously performs background replacement and illumination harmonization by concatenating the input noise with a foreground condition before feeding it into a diffusion model trained with a light transport consistency objective to ensure a coherent composite.

\noindent\textbf{Relighting in Video.}
With the maturation of image relighting techniques, researchers have begun to focus on the task of video relighting~\cite{guo2019relightables, zhang2021neural, fang2025relightvid, zhou2025light, gao2025anyportal}. Training-based methods such as IllumiCraft~\cite{lin2025illumicraft}, UniRelight~\cite{he2025unirelight}, RelightVid~\cite{fang2025relightvid}, Lumen~\cite{zeng2025lumen}, Lux Post Facto~\cite{mei2025lux}, and TC-Light~\cite{liu2025tc} perform video relighting by constructing datasets and training video-conditional diffusion models. However, collecting large-scale high-quality paired video relighting datasets is difficult \ysold{and requires high costs for training. 
Moreover, video models struggle to balance between the lighting conditions and the video foreground conditions. 
As a result, the trained models tend to produce lighting results with insufficient richness and diversity in illumination, and lacking fidelity under complex conditions.}

\begin{figure*}[htbp]
    \centering
    \includegraphics[width=\textwidth]{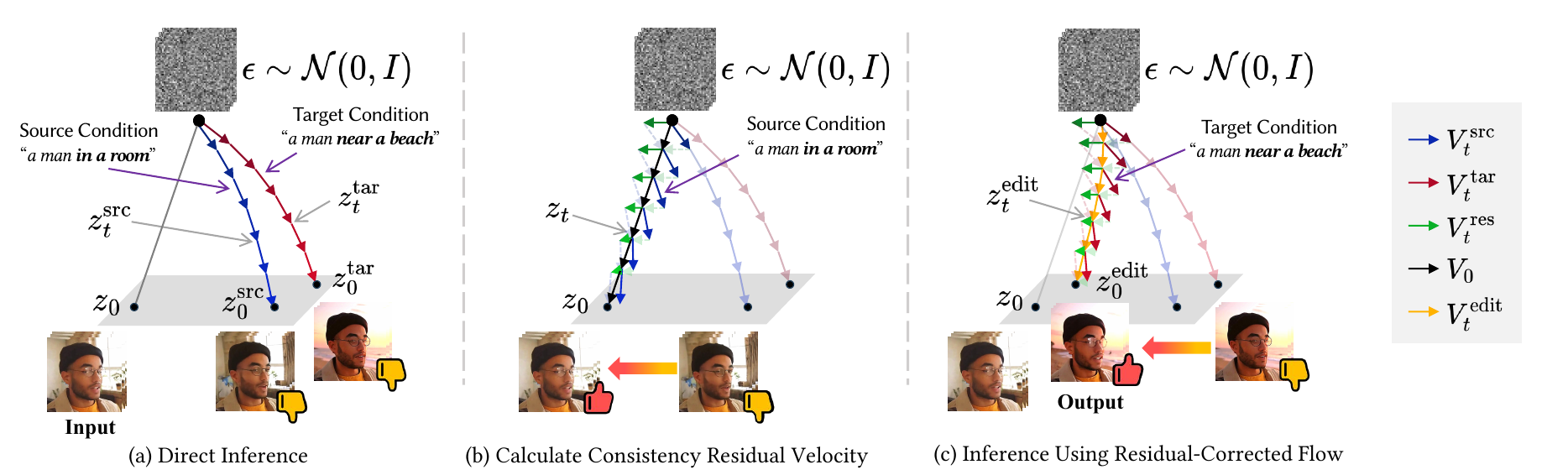}\vspace{-2mm}
    \caption{\textbf{Illustration of the proposed Residual-Corrected Flow.}
    (a) The \textbf{Naive Edit Flow} builds denoising trajectories under source and target conditions using the same noise \( \epsilon \). When applied to a real input video \( z_0 \), the mismatch between \( z_0^{\text{src}} \) and \( z_0 \) violates Stability under Identity. 
    (b) \textbf{Consistency Residual Velocity} \( V_t^{\text{res}} \) is constructed as the difference between the ideal restoration path \( V_0 \) and the predicted source flow \( V_t^{\text{src}} \), aligning the generated \( z_0^{\text{src}} \) with \( z_0 \). 
    (c) \textbf{Residual-Corrected Flow} combines \( V_t^{\text{tar}} \) and \( V_t^{\text{res}} \) to perform reliable video relighting that preserves identity consistency (\eg, mouth shape, glasses reflection) while enabling directional condition change. \ysold{The purple arrows indicate the condition to guide the velocity calculation. For simplicity, we omit the reference frame and structural conditions.}
    }
    \label{fig:flow1}
    \vspace{-3mm}
\end{figure*}

Some works pursue training-free approaches for efficiency. AnyPortal~\cite{gao2025anyportal} and Light-A-Video~\cite{zhou2025light} both adopt training-free (zero-shot) strategies, combining the capabilities of IC-Light and video diffusion generation models during inference. AnyPortal introduces cross-frame attention in IC-Light and applies a Refinement Projection Algorithm in the video model to ensure consistency, while Light-A-Video uses Consistent Light Attention in IC-Light and employs a Progressive Light Fusion strategy in the video diffusion process to maintain lighting generation and consistency.
However, in these two training-free methods, IC-Light is still used to relight per frame individually, introducing frame discontinuities that cannot be fully corrected by the video model. In addition, due to the lack of sufficient control in the video model, the relighting results may exhibit inconsistencies in structure and motion compared to the original video. Overcomplicated pipelines also make these methods less inference-efficient. 



\ysold{Our method is also training-free. Rather than per-frame IC-Light processing, we adopt holistic relighting to achieve better temporal consistency. Moreover, by introducing the proposed Residual-Corrected Flow and High-Frequency Transfer, we better address structural and detail consistency. To achieve stronger conditional control, we propose a Decoupled Condition Design to enhance the model's ability to respond to given conditions. We further achieve higher efficiency by reusing residual predictions for acceleration.}

\section{Method}
\label{sec:method}

\subsection{Preliminary: Flow Models}
In flow-based generative models~\cite{kong2024hunyuanvideo, wan2025wan, liu2022flow, lipman2022flow}, data generation process is modeled as a learned continuous flow that gradually transforms a random noise into a data sample\footnote{For simplicity, we omit the details regarding the latent space. The video is mapped into the latent space through a VAE.}. 

Formally, let \( z_t \) denote the latent variable at timestep (or noise level) \( t \in [0, 1] \), where \( t = 1 \) corresponds to the starting point of the generation process with \( z_1 \sim \mathcal{N}(0, I) \), and \( t = 0 \) indicates the final clean data sample \( z_0 \). Let \( c \) represent the conditional information.  
The model \( F_\theta \) predicts a velocity (or flow) vector field \( V_t^c \) at each noise level $t$:
\begin{equation}
\label{eq:pre1}
V_t^c(z_t) = F_\theta(z_t, t, c).
\end{equation}
\( V_t^c \) therefore defines how the latent variable evolves continuously from noise to data over time.  
This generation process follows an ordinary differential equation (ODE) form:
\begin{equation}
\label{eq:pre2}
\frac{d z_t}{d t} = V_t^c(z_t).
\end{equation}
In practice, this continuous process is discretized into \( N + 1 \) steps \(\{t_0, t_1, \dots, t_N\}\), with \(0 = t_0 < t_1 < \dots < t_N = 1\). 
Starting from an initial latent \( z_{t_N} = z_1 \sim \mathcal{N}(0, I) \),
 the model iteratively applies the discrete update rule
\begin{equation}
\label{eq:pre3}
z_{t_{i-1}} = z_{t_i} + (t_i - t_{i-1}) \, V_{t_i}^c(z_{t_i}),
\end{equation}
for \( N \) steps from \( t_N \) to \( t_0 \), progressively refining the latent until it reaches \( z_{t_0} = z_0 \) that lies in the data manifold.  

\subsection{Condition Consistency}

\ysold{\textbf{Condition Consistency} describes a fundamental property of an ideal editing model: changes in the output are driven by changes in the input conditions. It suggests two core principles: \textbf{1)} When the editing condition changes, a corresponding and perceivable change should manifest in the output; \textbf{2)} When the editing condition remains identical, the output should be a faithful replica of the input. 
This principle is critically important in the specific context of video relighting.
Specifically, it requires that 
\begin{itemize}
    \item Directional Change: The generated video should differ from the input only in lighting if the conditions differ only in lighting. All other aspects, including object structure and motion dynamics, must be perfectly preserved.
    \item Stability under Identity: In the extreme case where the target and source conditions are identical, an ideal model must reproduce the source video exactly.
\end{itemize}
It guarantees a faithful mapping of the condition signal, which neither omits required changes nor introduces extraneous ones, thereby ensuring predictable outcomes and high visual fidelity.
To this end, we aim to design a reliable video relighting model enforced with Condition Consistency.
}





\subsection{Residual-Corrected Flow with Decoupled Condition for Reliable Video Relighting} \label{sec:DRflow}

\ysold{\textbf{Naive Edit Flow.} We begin with a simple edit flow. We first sample a fixed Gaussian noise \( \epsilon \sim \mathcal{N}(0, I)\) and build two denoising trajectories over \( \epsilon \) under the source condition ``src'' (\eg, a prompt describing the original video) and the target condition ``tar'' (\eg, a prompt describing the target edited video) following Eq.~(\ref{eq:pre1})-(\ref{eq:pre3}). This results in two distinct flows denoted as \( V_{t}^{\text{src}} \) and \( V_{t}^{\text{tar}} \), respectively. The noise variable evolves along the flows from \( t = 1 \) to \( 0 \), producing two outputs: \( z_0^{\text{src}} \) and \( z_0^{\text{tar}} \), as shown in Fig.~\ref{fig:flow1}(a). 
Since two flows share the same $\epsilon$, if ``tar''$=$``src'', then $z_0^{\text{tar}}=z_0^{\text{src}}$. Clearly, the editing from a synthetic video \( z_0^{\text{src}} \) to \( z_0^{\text{tar}} \) satisfies the property of ``Stability under Identity''. 
However, issues arise when applying this edit flow to a real input video $z_0$. Due to inherent randomness, model capability limitations, and insufficient information in condition ``src'', the generated video \( z_0^{\text{src}} \) often does not exactly match \( z_0 \).
This violates Stability under Identity: when ``tar''$=$``src'', the output \( z_0^{\text{tar}}\neq z_0 \). 
}


\noindent\textbf{Residual-Corrected Flow.} \ysold{Based on the above analysis, our key idea is to simultaneously adjust both \( z_0^{\text{src}} \) and \( z_0^{\text{tar}} \), pulling \( z_0^{\text{src}} \) to be identical to \( z_0 \), so that the adjusted \( z_0^{\text{tar}} \), which we denote as \( z_0^{\text{edit}} \), can reflect directional change to \( z_0 \).}
First, we construct a restoration path from the noise \( \epsilon \) to the source video \( z_0 \).  
This path corresponds to a velocity defined as:
\begin{equation}
V_0 = \frac{z_0 - \epsilon}{1 - 0}.
\end{equation}
If the denoising process strictly follows this velocity, the noise \( \epsilon \) will be exactly transformed back into \( z_0 \), and the intermediate result \( z_{t} \) at timestep \( t \) satisfies
\begin{equation}
z_{t} = (1 - t) z_0 +  t \epsilon.
\end{equation}

However, in practice, the model predicts a velocity \( V_{t}^{\text{src}}(z_{t})\neq V_0  \) at \( z_{t} \).  To align this predicted flow with the ideal restoration path, we build a \textbf{Consistency Residual Velocity} \( V_{t}^{\text{res}} \) at each timestep, defined as:
\begin{equation}
V_{t}^{\text{res}}(z_{t}) = V_0 - V_{t}^{\text{src}} (z_{t}).
\end{equation}
After obtaining $V_{t}^{\text{res}}$ shown in Fig.~\ref{fig:flow1}(b), we can perform denoising starting from \( \epsilon \) using the combined flow \( V_{t}^{\text{src}} (z_{t}) + V_{t}^{\text{res}}(z_{t}) = V_0 \).  Following this adjusted flow trajectory ensures an accurate reconstruction of \( z_0 \) from $\epsilon$.

To achieve video transfer under the target condition, we introduce a \textbf{Residual-Corrected Flow} based on the Consistency Residual Velocity.  
Specifically, given the model-predicted flow  \( V_{t}^{\text{tar}} \) under the target condition and \( V_{t}^{\text{res}} \), we define the final Residual-Corrected Flow as:
\begin{equation}
V_{t}^{\text{edit}}(z^{\text{edit}}_{t}) = V_{t}^{\text{tar}}(z^{\text{edit}}_{t}) + V_{t}^{\text{res}}(z_{t}).
\end{equation}
Starting from the same initial Gaussian noise \( z^{\text{edit}}_1=\epsilon \), we then perform the denoising process following \( V_{t}^{\text{edit}} \) from \( t = 1 \) to \( 0 \), shown in Fig.~\ref{fig:flow1}(c).  
The final result of this process is our target relit video \( z_0^{\text{edit}} \).

\ysold{It is not hard to prove that when ``tar''$=$``src'', then $V_{t}^{\text{edit}}=V_0$ and \( z_0^{\text{tar}}= z_0 \), thereby satisfying Stability under Identity. Our next step is to enforce ``Directional Change'' property to achieve an overall Condition Consistency.}


\begin{figure}[t]
    \centering
    \includegraphics[width=\linewidth]{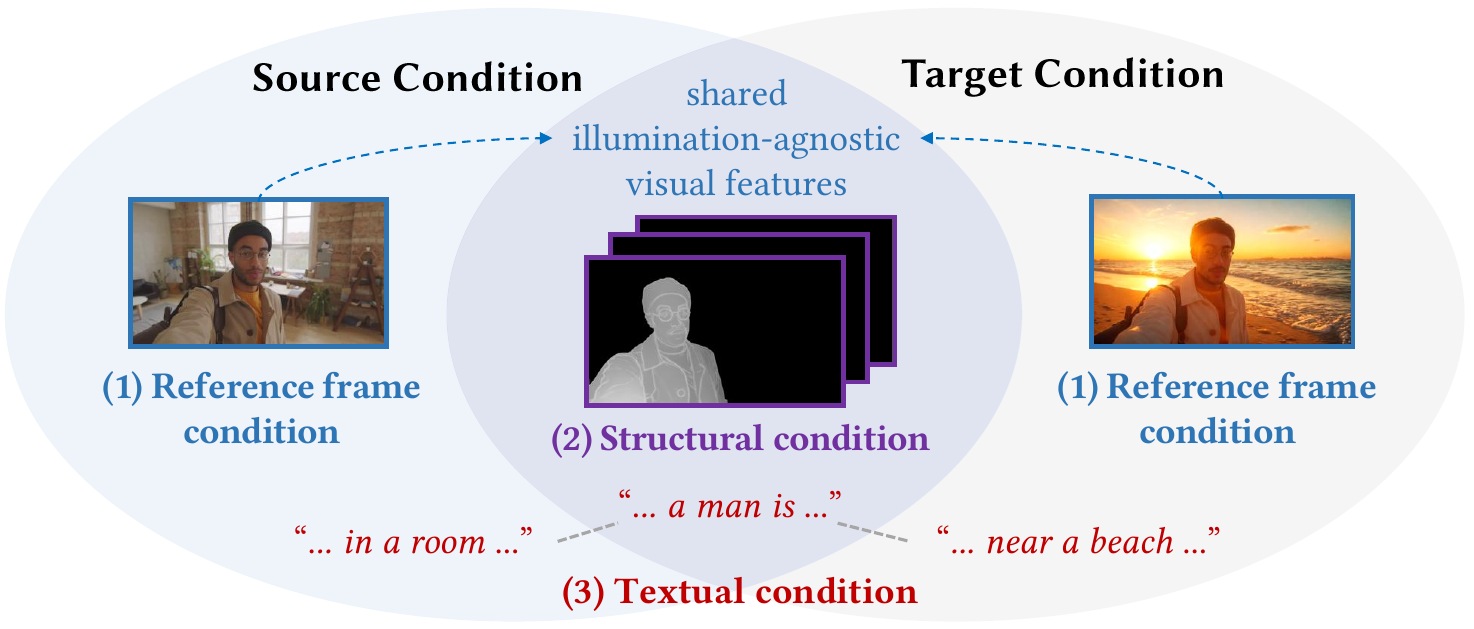}
    \caption{\textbf{Decoupled Condition Design.} The source and target conditions share identical illumination-agnostic information, differing only in their illumination-specific information.}
    \label{fig:cond}
    \vspace{-5mm}
\end{figure}

\noindent\textbf{Decoupled Condition Design}. \ysold{To strengthen  Directional Change}, we introduce a Decoupled Condition Design. Our condition input is carefully crafted using both illumination-specific and agnostic features to enable the model to more effectively identify what and where to make edits. As shown in Fig.~\ref{fig:cond}, it comprises:
\begin{itemize}
    \item \textbf{Reference frame condition.} \ysold{We use an I2V model as \( F_\theta \) to accept a reference frame. For the source condition, the reference frame is the first frame of the input video. For the target condition, it is the first frame edited by the powerful image relighting model, IC-Light~\cite{zhang2025scaling}, to provide a robust visual anchor that guarantees both illumination naturalness and spatial fidelity.}
    \item \textbf{Structural condition.} \ysold{We employ a pretrained ControlNet~\cite{zhang2023adding, videoxfun} on a weighted fusion of the depth and edge maps extracted from the input video as the illumination-agnostic structural condition. It is shared by both source and target conditions, ensuring the preservation of structural content during editing.}
    \item \textbf{Textual condition.} \ysold{We use illumination-specific textual prompts, where the source and target prompts differ only in their descriptions of background and lighting, while sharing the same foreground content.} 
\end{itemize}
This decomposition provides a stable, sufficient, and disentangled guidance signal to enforce Condition Consistency.

\subsection{High-Frequency Transfer} \label{sec:HF-transfer}

To further enhance the detail consistency between the generated target video and the original source video, we aim to inject a portion of high-frequency information from the source video into the generation process of the target video.  

Specifically, let \( \text{HF}(X) \) and \( \text{LF}(X) \) denote the high-frequency and low-frequency components of the video \( X \) obtained by Fourier decomposition satisfying 
$
X = \text{HF}(X) + \text{LF}(X).
$
Then, during each generation step of the target video \( z_{t}^{\text{edit}} \), we perform the following replacement:
\begin{equation}
z_{t}^{\text{edit}} \gets \text{LF}(z_{t}^{\text{edit}})
+ \lambda \cdot \text{HF}(z_{t})
+ (1 - \lambda) \cdot \text{HF}(z_{t}^{\text{edit}})
\end{equation}
where $z_{t}$ is the source video at step $t$ and \( \lambda \) controls the proportion of high-frequency information injected. 
A larger \( \lambda \) strengthens the preservation of fine details from the source video but may limit the model's ability to fully adapt to the target illumination condition; conversely, a smaller \( \lambda \) allows more flexibility in relighting but may reduce structural and textural consistency.

Note that the injection maintains Stability under Identity since when reconstructing \( z_0 \), transferring high-frequency from \( z_{t} \) to \( z_{t} \) itself does not alter itself.

\subsection{High-Quality Background Generation with Masking Mechanism}

\ysold{Residual-Corrected Flow and High-Frequency Transfer may introduce unwanted background details from source video into the result. 
To mitigate this conflict during new background generation,} we use the mask to disentangle the foreground and background regions.
Let \( M \) denote the mask of the foreground region extracted from the original video. Then, the Masked Residual-Corrected Flow is
\begin{equation}
V_{t}^{\text{edit}}(z_t^{\text{edit}}) = V_{t}^{\text{tar}}(z_t^{\text{edit}}) + M \cdot V_{t}^{\text{res}}(z_t).
\end{equation}
\ysold{Similarly, we apply the Masked High-Frequency Transfer only to the foreground region:}
\begin{equation}
z_{t}^{\text{edit}} \gets \text{LF}(z_{t}^{\text{edit}})
+  \lambda M \cdot \text{HF}(z_{t})
+ (1 -  \lambda M) \cdot \text{HF}(z_{t}^{\text{edit}}).
\end{equation}
\ysold{The structural condition of our Decoupled Condition Design is modified accordingly.} We utilize a combination of the previously extracted features (\ie, depth and edge), but only for the foreground region, leaving the background unconstrained to allow new content generation.

\ysold{Using this mask mechanism, we can easily preserve the foreground details while generating the background unperturbedly without any interference from the source video. Note that this mask mechanism permits certain regions to violate the Stability under Identity property, allowing for more flexible editing.
}


\subsection{Comparison to FlowEdit in Video Relighting} 
\label{Sec. ana}

\ysold{
Inversion and FlowEdit also aim for editing based on reconstruction. 
Although inversion~\cite{song2020denoising, wu2023latent, hertz2022prompt, mokady2023null} is commonly used in image generation models for editing, it remains both time-consuming and imprecise in video models. 
\ysold{The imprecision also violates the Stability under Identity, resulting in poor Condition Consistency. By comparison, our method is inversion-free, which is partially inspired by the recent approach, FlowEdit~\cite{kulikov2025flowedit}. It offers a novel path for image editing that satisfies Stability under Indentity without inversion. However, it has obvious limitations when applied to the video relighting task.
}
}

In FlowEdit, starting with the source image \( z^{\text{edit}}_1 = z_0 \), the target image \( z^{\text{edit}}_{t} \) evolves along the flow \( V^{\text{edit}}_{t} \):
\begin{align}
\label{eq:flowedit}
V^{\text{edit}}_{t}(z^{\text{edit}}_{t}) &= V^{\text{tar}}_{t}(z^{\text{pred}}_{t}) - V^{\text{src}}_{t}(z_{t}),\\
z_{t} &= (1 - t)z_0 + t\epsilon_t,\\
z^{\text{pred}}_{t} &= z_{t} + z^{\text{edit}}_{t} - z_0,
\end{align}
from \( t = 1 \) to \( 0 \), where \( \epsilon_t \sim  \mathcal{N}(0, I) \). 
\ysold{Our design differs from FlowEdit in two aspects: 1) FlowEdit samples $n$ different Gaussian noises at each timestep while we propose to use a single Gaussian noise $\epsilon$ throughout the whole editing process. 2) FlowEdit constructs an edit flow from the source video $z^{\text{edit}}_1=z_0$ to target result $z^{\text{edit}}_0$.
We rephrase this process and build a new generation flow from noise $z^{\text{edit}}_1=\epsilon$ to the target $z^{\text{edit}}_0$ by tracking $z^{\text{pred}}_{t}$ in FlowEdit, just as the standard diffusion denoising process.}

\begin{figure}[t]
    \centering
    \includegraphics[width=0.8\linewidth]{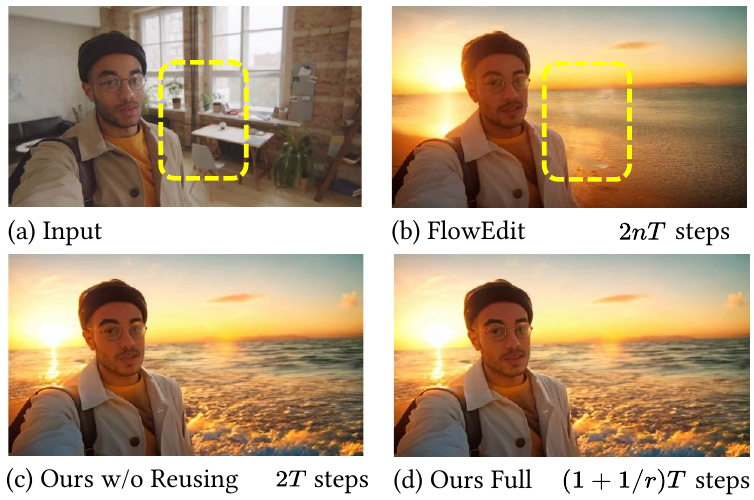}\vspace{-2mm}
    \caption{\textbf{Comparison with FlowEdit.} (a) Input video. (b) FlowEdit produces blurry outputs, and suffers from ghosting artifacts due to the interference from the original background (yellow region). (c)(d) Our residual reusing strategy effectively reduces the number of prediction steps with negligible quality degradation.}
    \label{fig:flowedit}
    \vspace{-5mm}
\end{figure}

\ysold{It can be proved that if FlowEdit samples a fixed noise across steps, it is theoretically equivalent to our method without masking mechanism. However, it is the above two straightforward changes that make a big difference, establishing key advantages of our approach over FlowEdit:}
\begin{itemize}
    \item \ysold{Clear Generation Ability.~To ensure stable editing, FlowEdit independently samples multiple Gaussian noises and averages \( V^{\text{edit}}_{t} \) at each timestep, which inevitably causes blur in the generated background as in Fig.~\ref{fig:flowedit} (b). Our fixed noise prevents blur, significantly improving the generation quality.
    \item Acceleration. Our method and FlowEdit require computing two velocities, doubling the denoising steps from $T$ to $2T$ \ysold{($2nT$ for FlowEdit if averaging $n\geq1$ velocities per step)}. The benefit we offer is that the proposed consistency residual velocity relies solely on the fixed noise and source condition, thus it is stable across steps and can be reused. Our experiments have shown that applying the same residual velocity every \( r \) steps has little impact on quality, while it decreases the total steps to \( (1 + 1/r)T \), as shown in Fig.~\ref{fig:flowedit} (c)(d). In contrast,  FlowEdit requires new random noises and target-condition denoising for each edit flow calculation, which prevents reusability.
    \item Spatial Controllability. The edit flow of FlowEdit starts from the source video, which cannot handle pure generation within specific regions. The background from the source video will interfere with the new background generation as in Fig.~\ref{fig:flowedit} (b). By comparison, our method, starting from pure noises, enables foreground masking for editing while leaving the background for pure generation. 
    \item Decoupled Condition. Additionally, we propose a Decoupled Condition Design to combine the reference frame and structural conditions, which helps the model achieve better Condition Consistency.}   
\end{itemize}

\section{Experiments}
\label{sec:exp}

\begin{figure*}[htbp]
    \centering
    \includegraphics[width=0.92\textwidth]{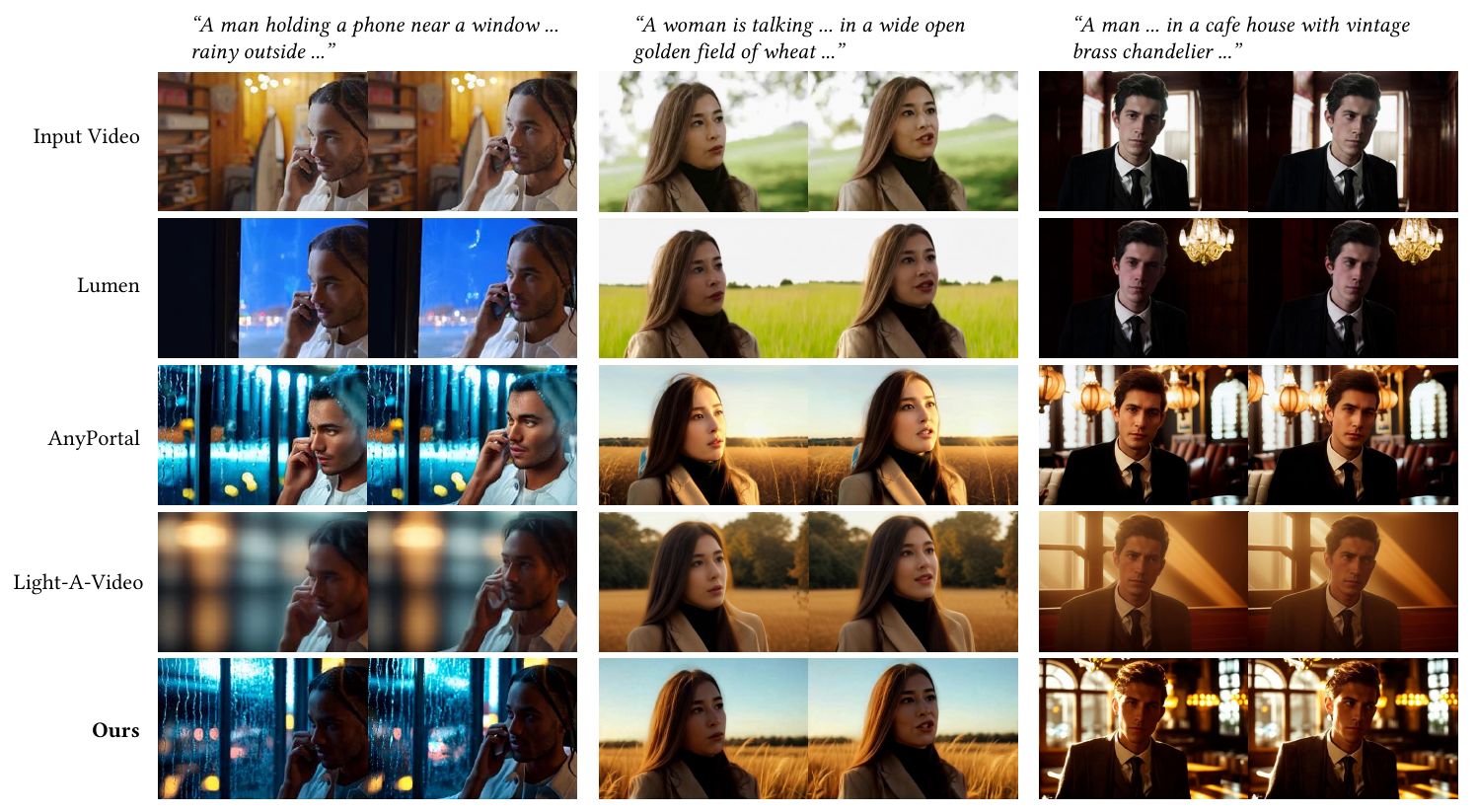}
    \vspace{-3mm}
    \caption{\textbf{Qualitative comparison.} The training-based Lumen exhibits insufficient lighting richness and  diversity, with foreground nearly unaltered. The training-free AnyPortal and Light-A-Video show poor structural fidelity and lighting quality. Our method not only maintains structural and detail consistency but also demonstrates high-quality background generation and rich relighting effects.}
    \label{fig:comp}
    \vspace{-1mm}
\end{figure*}

\subsection{Experimental Setup}

\textbf{Implementation details.} 
We implement our method based on Wan2.1~\cite{wan2025wan, videoxfun}, a widely used state-of-the-art open-sourced video diffusion model. 
For the foreground mask $M$, we use BiRefNet~\cite{zheng2024bilateral} to generate the mask for the first frame and MatAnyone~\cite{yang2025matanyone} to propagate it throughout the entire video. \ysold{We downsample $M$ to match the video shape in the latent space for masking operation.}
For the Decoupled Condition Design, the reference frame is generated by IC-Light~\cite{zhang2025scaling}, and the structural information is obtained by combining the HED, depth, and Canny maps~\cite{zhang2023adding}, followed by masking with $M$.
For Residual-Corrected Flow, we adopt $T=50$ timesteps for the generation process and reuse the Consistency Residual Velocity every $r=10$ steps, resulting in a total of $55$ steps. 
For High-Frequency Transfer, we set the frequency decomposition threshold to $0.8$ and the injection intensity to $\lambda = 0.5$. 
 The experiments are conducted on a single 80 GB NVIDIA A100 GPU. Representative results are illustrated in Fig.~\ref{fig:teaser} and Fig.~\ref{fig:more}.

\noindent\textbf{Baseline methods.}
We compare our method with recent video relighting approaches. Training-free methods such as AnyPortal~\cite{gao2025anyportal} and Light-A-Video~\cite{zhou2025light} apply the image relighting model IC-Light frame by frame and integrate the results into video generation models. Training-based methods like Lumen~\cite{zeng2025lumen} and TC-Light~\cite{liu2025tc} are trained on paired video datasets constructed for the relighting task. The basic information for these methods is summarized in Tab.~\ref{tab:method_basics}. For fairness, we compare only with methods that support new background generation, excluding RelightVid~\cite{fang2025relightvid} and Lux Post Facto~\cite{mei2025lux} since they are not open-sourced.

\begin{table}[t]
\centering
\caption{Summarization of video relighting methods.}\vspace{-2mm}
\label{tab:method_basics}
\footnotesize
\setlength{\tabcolsep}{2pt}
\begin{tabular}{lcccc}
\toprule
\textbf{Method} &
\makecell{\textbf{Training-}\\\textbf{Free}} &
\makecell{\textbf{Video}\\\textbf{Base}\\\textbf{Model}} &
\makecell{\textbf{Support}\\\textbf{Diverse}\\\textbf{Resolution}} &
\makecell{\textbf{Support}\\\textbf{Background}\\\textbf{Replacement}} \\
\midrule
AnyPortal & $\surd$ & CogVideoX & $\times$ & $\surd$ \\
Light-A-Video (A) & $\surd$ & AnimateDiff & $\surd$ & $\surd$ \\
Light-A-Video (C) & $\surd$ & CogVideoX & $\times$ & $\times$ \\
Light-A-Video (W) & $\surd$ & Wan2.1 & $\surd$ & $\times$ \\
Lumen & $\times$ & Wan2.1 & $\surd$ & $\surd$ \\
TC-Light & $\times$ & VidToMe & $\surd$ & $\times$ \\
\textbf{FlowPortal (Ours)} & $\surd$ & Wan2.1 & $\surd$ & $\surd$ \\
\bottomrule
\end{tabular}\vspace{-2mm}
\end{table}


\noindent\textbf{Metrics.}  
To comprehensively evaluate the quality of relighted videos, we assess our results from four perspectives. 
\begin{itemize}
    \item \textbf{Video–text alignment.} We employ CLIP-T \ysold{cosine} similarity~\cite{qi2023fatezero} to measure the alignment between generated frames and the target relighting prompts.
    \item \textbf{Temporal smoothness.} We use CLIP-I~\cite{qi2023fatezero}, \ysold{the CLIP-based cosine similarity between consecutive frames}, to evaluate temporal consistency in the generated videos.
    \item \textbf{Foreground consistency.} As existing metrics rarely address the consistency of primary subjects in relighting tasks, we propose a new evaluation protocol. Specifically, we first extract the foreground region~\cite{BiRefNet} of each output video, then compute (a) \textbf{Structural consistency} using SSIM between Canny edge maps~\cite{zhang2023adding}, (b) \textbf{Detail consistency} using PSNR between LoG responses~\cite{marr1980theory} of albedo predictions~\cite{careaga2024colorful}, (c) \textbf{Motion consistency} using SSIM between optical flow maps~\cite{teed2020raft}, and (d) \textbf{Identity consistency} by calculating facial ID similarity~\cite{deng2019arcface} for human-centric clips or CLIP-I similarity for others.
    \item \textbf{User preference.} We invite 23 participants to select the best result among four methods based on four criteria: (a) User-Pmt (relevance to the prompt), (b) User-Tmp (temporal coherence), (c) User-Fg (preservation of foreground details and motion), and (d) User-Lit (quality and harmonization of relighting on the foreground). 
\end{itemize}


\subsection{Comparison to State-of-the-Art Methods}

\textbf{Qualitative results.} Figure~\ref{fig:comp} shows the qualitative comparison between FlowPortal and other methods. The training-based method Lumen fails to accurately preserve detail fidelity and also exhibits poor lighting alteration and richness. The training-free methods AnyPortal and Light-A-Video produce poor foreground consistency, background quality, and lighting harmonization.

\begin{table*}[h]
\centering
\caption{Quantitative and user study results.}\vspace{-2mm}
\label{tab:quantitative_user}
\footnotesize
\setlength{\tabcolsep}{3pt}
\resizebox{\linewidth}{!}{
\begin{tabular}{l|ccccccc|cccc}
\toprule
\textbf{Method} &
\makecell{\textbf{CLIP-T}} &
\makecell{\textbf{CLIP-I}} &
\makecell{\textbf{Structural}\\\textbf{Consistency}} &
\makecell{\textbf{Motion}\\\textbf{Consistency}} &
\makecell{\textbf{Detail}\\\textbf{Consistency}} &
\makecell{\textbf{Identity}\\\textbf{Consistency}} &
\makecell{\textbf{}} &
\makecell{\textbf{User-Pmt}} &
\makecell{\textbf{User-Tmp}} &
\makecell{\textbf{User-Fg}} &
\makecell{\textbf{User-Lit}} \\
\midrule
AnyPortal  & \underline{0.3196} & \underline{0.9817} & 0.8530 & 0.8876 & 40.4853 & 0.4310 & & 15.3 & 9.9 & 8.9 & 12.1 \\
Light-A-Video (A) & 0.2956 & 0.9684 & 0.8580 & 0.8869 & \underline{40.8727} & 0.5076 & & 4.7 & 3.7 & 3.4 & 9.4 \\
Lumen  & 0.3055 & 0.9746 & \textbf{0.8809} & \underline{0.8914} & 40.4193 & \textbf{0.7392} & & \underline{22.7} & \underline{28.1} & \underline{35.5} & \underline{24.4} \\
\textbf{FlowPortal (Ours)} & \textbf{0.3271} & \textbf{0.9828} & \underline{0.8804} & \textbf{0.8944} & \textbf{41.2044} & \underline{0.7328} & & \textbf{57.4} & \textbf{58.4} & \textbf{52.2} & \textbf{54.2} \\
\bottomrule
\end{tabular}}
\vspace{-3mm}
\end{table*}

\begin{figure*}[t]
    \centering
    \includegraphics[width=0.89\linewidth]{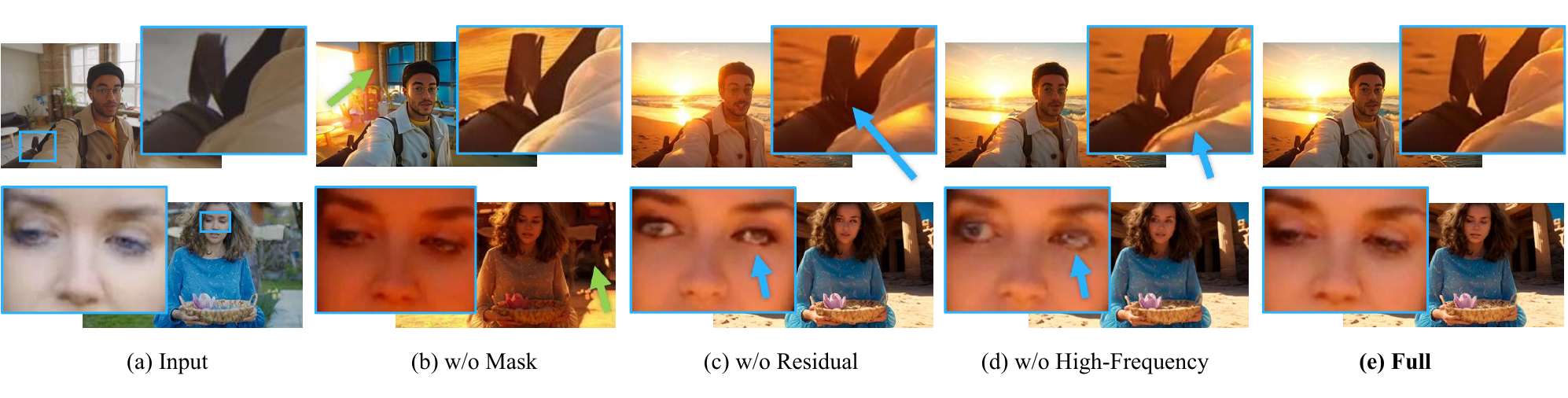}\vspace{-4mm}
    \caption{\textbf{Ablation Study. }Without mask, background cannot be properly generated in (b). Enlarged local blue regions illustrate the structural inconsistency if (c) removing Residual-Corrected Flow and detail inconsistency if (d) removing High-Frequency Transfer.}\vspace{-2mm}
    \label{fig:abla}
\end{figure*}


\noindent\textbf{Quantitative results.} 
We construct a test set consisting of 69 pairs of real-world video clips and corresponding relighting prompts with broad diversity, including 54 human-centric videos, 8 animal scenes, and 7 other objects for objective evaluation. Table~\ref{tab:quantitative_user} shows the results among different methods. Our method achieves the best video-text alignment, temporal smoothness, motion consistency, and detail consistency. The training-based method Lumen introduces less variation in lighting conditions with foreground nearly unaltered, which achieves the best detail consistency and identity consistency. Our method ranks the second with negligible consistency value drops, but achieves significant lighting adjustment, indicating our method's high Condition Consistency.
For subjective evaluation, 
Table~\ref{tab:quantitative_user} shows the user preference scores averaged over 17 randomly selected results and 24 participants. Our method achieves the best 
overall user preference.

\noindent\textbf{Running time.}~We report the running time of training-free methods on an 80 GB NVIDIA A100 GPU. AnyPortal and Light-A-Video require 20–30 minutes per video due to complex pipelines. In contrast, our method only takes 3–5 minutes across different resolutions roughly equivalent to the direct inference time of a single video diffusion model.

\subsection{Ablation Study}

We conduct ablation studies to evaluate the effectiveness of Residual-Corrected Flow, High-Frequency Transfer, and the masking mechanism. The qualitative results are shown in Fig.~\ref{fig:abla}, and the quantitative results are presented in Tab.~\ref{tab:abla}.
\begin{itemize}
    \item \textbf{Masking mechanism.} Inference without masking the Consistency Residual Velocity and High-Frequency Transfer causes interference with the generation of the new background, resulting in an unchanged background structure and poor prompt relevance.
    \item \textbf{Residual-Corrected Flow.} Performing direct inference instead of using the Residual-Corrected Flow leads to severe structural incoherence.
    \item \textbf{High-Frequency Transfer.} The texture-level details fail to be preserved without High-Frequency Transfer due to inaccuracy in structural information and the model's imperfect controllability under these conditions.
\end{itemize}

\begin{table}[t]
\centering
\caption{Quantitative ablation study.}\vspace{-2mm}
\label{tab:abla}
\footnotesize
\setlength{\tabcolsep}{5pt}
\begin{tabular}{l|cccc}
\toprule
\textbf{Metric} &
\makecell{\textbf{w/o}\\\textbf{Mask}} &
\makecell{\textbf{w/o}\\\textbf{Residual}} &
\makecell{\textbf{w/o High-}\\\textbf{Frequency}} &
\textbf{Full} \\
\midrule
CLIP-T & 0.2809 & \textbf{0.3310} & \underline{0.3290} & 0.3271 \\
CLIP-I & 0.9792 & \underline{0.9825} & 0.9798 & \textbf{0.9828} \\
Structural Cons. & 0.8649 & 0.8516 & \underline{0.8688} & \textbf{0.8804} \\
Motion Cons. & 0.8923 & \underline{0.8933} & 0.8882 & \textbf{0.8944} \\
Detail Cons. & 40.4353 & 38.5027 & \underline{40.4852} & \textbf{41.2044} \\
Identity Cons. & \underline{0.7123} & 0.4153 & 0.5527 & \textbf{0.7328} \\
\bottomrule
\end{tabular}\vspace{-3mm}
\end{table}

\begin{figure*}[htbp]
    \centering
    \includegraphics[width=0.92\textwidth]{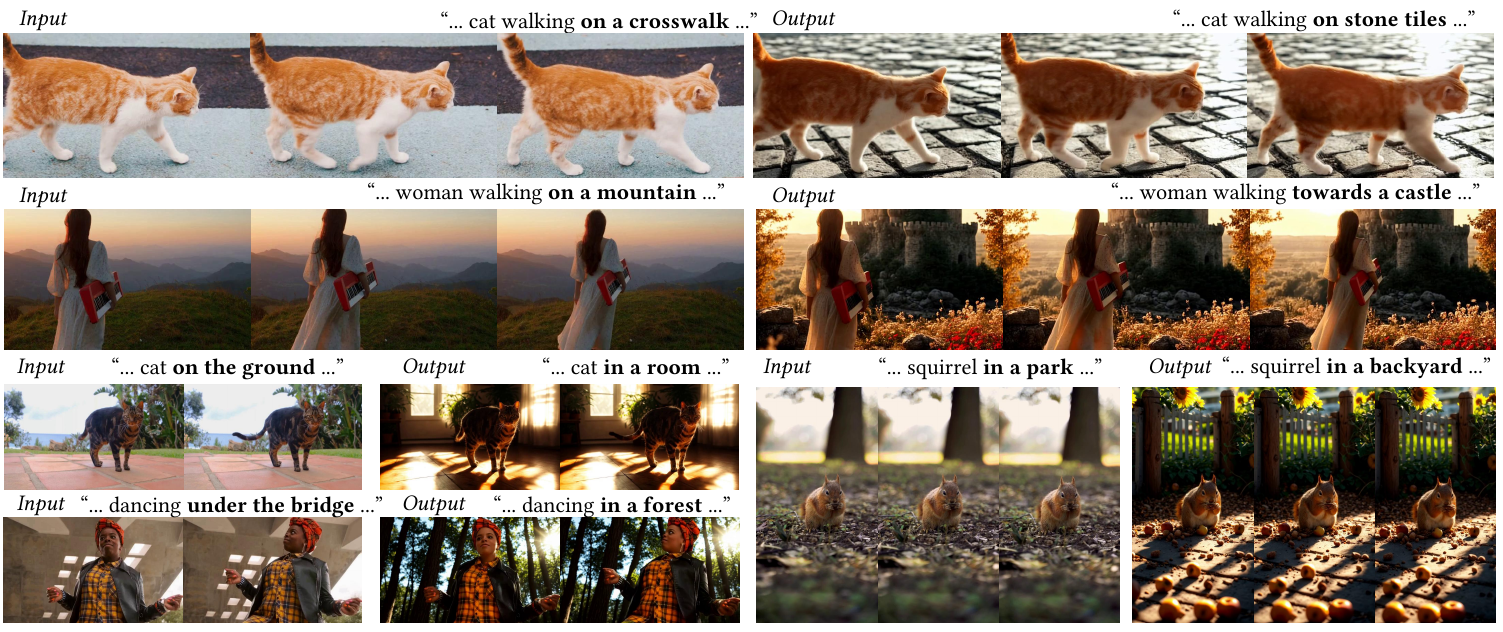}
    \vspace{-2mm}
    \caption{\textbf{More visual results.} Our method can generate realistic backlighting effects that simulate light penetrating through clothing, as well as clear shadows cast by objects onto the background under strong lighting conditions. See our project page: https://gaowenshuo.github.io/FlowPortalProject/ for video demonstration.}
    \label{fig:more}
    \vspace{-3mm}
\end{figure*}

\begin{figure}[t]
    \centering
    \vspace{-2mm}
    \includegraphics[width=0.85\linewidth]{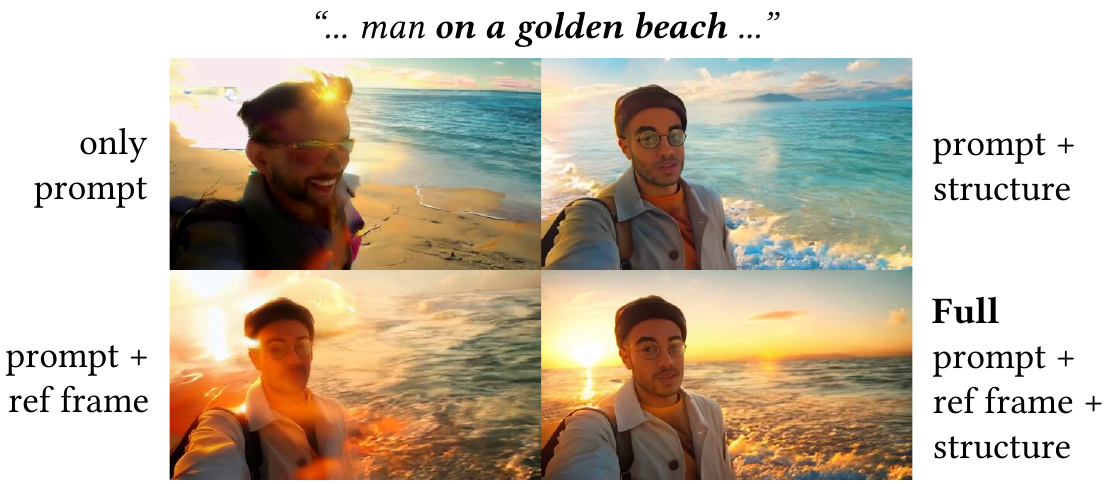}  \vspace{-2mm}
    \caption{\textbf{Ablation on Decoupled Condition Design.} The absence of structural information or the reference frame leads to degraded structural fidelity and unnatural illumination.}
    \label{fig:dcd}
    \vspace{-7mm}
\end{figure}

\noindent\textbf{Decoupled Condition Design.}
\ysold{We study the effect of our Decoupled Condition Design in Fig.~\ref{fig:dcd}.
It can be seen that when using only text conditions, the foreground character fails to be reconstructed properly. Adding structural conditions maintains consistent character structure but fails to produce natural golden lighting. Incorporating the reference frame results in stronger and prompt-consistent lighting, but leads to incomplete character structure. Only when all conditions are utilized, does our method achieve the most natural results in both structure and lighting.}

\section{Conclusion and Discussion}

In this paper, we propose FlowPortal, a novel training-free framework for efficient video relighting and background replacement. We introduce a novel Residual-Corrected Flow mechanism with Decoupled Condition Design that enforces Condition Consistency. 
A High-Frequency Transfer module is designed to enhance the detail fidelity and a masking mechanism is applied to isolate background regions for high-quality generation. 
Both qualitative and quantitative experiments demonstrate that our method achieves superior background generation quality, foreground consistency, and lighting realism. 

Although our method performs well in most cases, it may still be constrained by the generative capability of the underlying base models (\ie, IC-Light for relighting the reference frame and Wan2.1 for background generation).

It is evident that our Residual-Corrected Flow framework is not exclusively designed for relighting.~A potential future direction could be the exploration of extending it to various video editing tasks, including video stylization, colorization, object manipulation, human editing, and even more general editing tasks.~Further algorithmic improvements on the Residual-Corrected Flow, reconstruction rather than relying on direct directional addition, would be another promising future direction. 

{
    \small
    \bibliographystyle{ieeenat_fullname}
    \bibliography{main}

\begin{thebibliography}{34}
\providecommand{\natexlab}[1]{#1}
\providecommand{\url}[1]{\texttt{#1}}
\expandafter\ifx\csname urlstyle\endcsname\relax
  \providecommand{\doi}[1]{doi: #1}\else
  \providecommand{\doi}{doi: \begingroup \urlstyle{rm}\Url}\fi

\bibitem[{AIGC-Apps}(2025)]{videoxfun}
{AIGC-Apps}.
\newblock Videox-fun: Github repository, 2025.
\newblock GitHub repository: github.com/aigc-apps/VideoX-Fun.

\bibitem[Careaga and Aksoy(2024)]{careaga2024colorful}
Chris Careaga and Ya{\u{g}}{\i}z Aksoy.
\newblock Colorful diffuse intrinsic image decomposition in the wild.
\newblock \emph{ACM TOG}, 43\penalty0 (6):\penalty0 1--12, 2024.

\bibitem[Deng et~al.(2019)Deng, Guo, Xue, and Zafeiriou]{deng2019arcface}
Jiankang Deng, Jia Guo, Niannan Xue, and Stefanos Zafeiriou.
\newblock Arcface: Additive angular margin loss for deep face recognition.
\newblock In \emph{CVPR}, pages 4690--4699, 2019.

\bibitem[Fang et~al.(2025)Fang, Sun, Zhang, Wu, Xu, Zhang, Wang, Wetzstein, and Lin]{fang2025relightvid}
Ye Fang, Zeyi Sun, Shangzhan Zhang, Tong Wu, Yinghao Xu, Pan Zhang, Jiaqi Wang, Gordon Wetzstein, and Dahua Lin.
\newblock Relightvid: Temporal-consistent diffusion model for video relighting.
\newblock \emph{arXiv preprint arXiv:2501.16330}, 2025.

\bibitem[Gao et~al.(2025)Gao, Lan, and Yang]{gao2025anyportal}
Wenshuo Gao, Xicheng Lan, and Shuai Yang.
\newblock Anyportal: Zero-shot consistent video background replacement.
\newblock In \emph{ICCV}, pages 18990--18999, 2025.

\bibitem[Guo et~al.(2019)Guo, Lincoln, Davidson, Busch, Yu, Whalen, Harvey, Orts-Escolano, Pandey, Dourgarian, et~al.]{guo2019relightables}
Kaiwen Guo, Peter Lincoln, Philip Davidson, Jay Busch, Xueming Yu, Matt Whalen, Geoff Harvey, Sergio Orts-Escolano, Rohit Pandey, Jason Dourgarian, et~al.
\newblock The relightables: Volumetric performance capture of humans with realistic relighting.
\newblock \emph{ACM TOG}, 38\penalty0 (6):\penalty0 1--19, 2019.

\bibitem[He et~al.(2025)He, Liang, Munkberg, Hasselgren, Vijaykumar, Keller, Fidler, Gilitschenski, Gojcic, and Wang]{he2025unirelight}
Kai He, Ruofan Liang, Jacob Munkberg, Jon Hasselgren, Nandita Vijaykumar, Alexander Keller, Sanja Fidler, Igor Gilitschenski, Zan Gojcic, and Zian Wang.
\newblock Unirelight: Learning joint decomposition and synthesis for video relighting.
\newblock \emph{arXiv preprint arXiv:2506.15673}, 2025.

\bibitem[Hertz et~al.(2022)Hertz, Mokady, Tenenbaum, Aberman, Pritch, and Cohen-Or]{hertz2022prompt}
Amir Hertz, Ron Mokady, Jay Tenenbaum, Kfir Aberman, Yael Pritch, and Daniel Cohen-Or.
\newblock Prompt-to-prompt image editing with cross-attention control.
\newblock In \emph{ICLR}, pages 1--12, 2022.

\bibitem[Kim et~al.(2024)Kim, Jang, Yoon, Lee, Na, and Woo]{kim2024switchlight}
Hoon Kim, Minje Jang, Wonjun Yoon, Jisoo Lee, Donghyun Na, and Sanghyun Woo.
\newblock Switchlight: Co-design of physics-driven architecture and pre-training framework for human portrait relighting.
\newblock In \emph{CVPR}, pages 25096--25106, 2024.

\bibitem[Kong et~al.(2024)Kong, Tian, Zhang, Min, Dai, Zhou, Xiong, Li, Wu, Zhang, et~al.]{kong2024hunyuanvideo}
Weijie Kong, Qi Tian, Zijian Zhang, Rox Min, Zuozhuo Dai, Jin Zhou, Jiangfeng Xiong, Xin Li, Bo Wu, Jianwei Zhang, et~al.
\newblock Hunyuanvideo: A systematic framework for large video generative models.
\newblock \emph{arXiv preprint arXiv:2412.03603}, 2024.

\bibitem[Kulikov et~al.(2025)Kulikov, Kleiner, Huberman-Spiegelglas, and Michaeli]{kulikov2025flowedit}
Vladimir Kulikov, Matan Kleiner, Inbar Huberman-Spiegelglas, and Tomer Michaeli.
\newblock Flowedit: Inversion-free text-based editing using pre-trained flow models.
\newblock In \emph{ICCV}, pages 19721--19730, 2025.

\bibitem[Lin et~al.(2025)Lin, Chen, Tsai, Clark, and Yang]{lin2025illumicraft}
Yuanze Lin, Yi-Wen Chen, Yi-Hsuan Tsai, Ronald Clark, and Ming-Hsuan Yang.
\newblock Illumicraft: Unified geometry and illumination diffusion for controllable video generation.
\newblock \emph{arXiv preprint arXiv:2506.03150}, 2025.

\bibitem[Lipman et~al.(2022)Lipman, Chen, Ben-Hamu, Nickel, and Le]{lipman2022flow}
Yaron Lipman, Ricky~TQ Chen, Heli Ben-Hamu, Maximilian Nickel, and Matt Le.
\newblock Flow matching for generative modeling.
\newblock \emph{arXiv preprint arXiv:2210.02747}, 2022.

\bibitem[Liu et~al.(2022)Liu, Gong, and Liu]{liu2022flow}
Xingchao Liu, Chengyue Gong, and Qiang Liu.
\newblock Flow straight and fast: Learning to generate and transfer data with rectified flow.
\newblock \emph{arXiv preprint arXiv:2209.03003}, 2022.

\bibitem[Liu et~al.(2025)Liu, Luo, Tang, Li, Yang, Ning, Fan, Peng, and Zhang]{liu2025tc}
Yang Liu, Chuanchen Luo, Zimo Tang, Yingyan Li, Yuran Yang, Yuanyong Ning, Lue Fan, Junran Peng, and Zhaoxiang Zhang.
\newblock Tc-light: Temporally consistent relighting for dynamic long videos.
\newblock \emph{arXiv preprint arXiv:2506.18904}, 2025.

\bibitem[Marr and Hildreth(1980)]{marr1980theory}
David Marr and Ellen Hildreth.
\newblock Theory of edge detection.
\newblock \emph{Proceedings of the Royal Society of London. Series B. Biological Sciences}, 207\penalty0 (1167):\penalty0 187--217, 1980.

\bibitem[Mei et~al.(2025)Mei, He, Ma, Philip, Xian, George, Yu, Dedic, Ta{\c{s}}el, Yu, et~al.]{mei2025lux}
Yiqun Mei, Mingming He, Li Ma, Julien Philip, Wenqi Xian, David~M George, Xueming Yu, Gabriel Dedic, Ahmet~Levent Ta{\c{s}}el, Ning Yu, et~al.
\newblock Lux post facto: Learning portrait performance relighting with conditional video diffusion and a hybrid dataset.
\newblock In \emph{CVPR}, pages 5510--5522, 2025.

\bibitem[Mokady et~al.(2023)Mokady, Hertz, Aberman, Pritch, and Cohen-Or]{mokady2023null}
Ron Mokady, Amir Hertz, Kfir Aberman, Yael Pritch, and Daniel Cohen-Or.
\newblock Null-text inversion for editing real images using guided diffusion models.
\newblock In \emph{CVPR}, pages 6038--6047, 2023.

\bibitem[Pandey et~al.(2021)Pandey, Orts-Escolano, Legendre, Haene, Bouaziz, Rhemann, Debevec, and Fanello]{pandey2021total}
Rohit Pandey, Sergio Orts-Escolano, Chloe Legendre, Christian Haene, Sofien Bouaziz, Christoph Rhemann, Paul~E Debevec, and Sean~Ryan Fanello.
\newblock Total relighting: learning to relight portraits for background replacement.
\newblock \emph{ACM TOG}, 40\penalty0 (4):\penalty0 43--1, 2021.

\bibitem[Qi et~al.(2023)Qi, Cun, Zhang, Lei, Wang, Shan, and Chen]{qi2023fatezero}
Chenyang Qi, Xiaodong Cun, Yong Zhang, Chenyang Lei, Xintao Wang, Ying Shan, and Qifeng Chen.
\newblock Fatezero: Fusing attentions for zero-shot text-based video editing.
\newblock In \emph{ICCV}, pages 15932--15942, 2023.

\bibitem[Ren et~al.(2024)Ren, Xiong, Yoon, Shu, Zhang, Jung, Gerig, and Zhang]{ren2024relightful}
Mengwei Ren, Wei Xiong, Jae~Shin Yoon, Zhixin Shu, Jianming Zhang, HyunJoon Jung, Guido Gerig, and He Zhang.
\newblock Relightful harmonization: Lighting-aware portrait background replacement.
\newblock In \emph{CVPR}, pages 6452--6462, 2024.

\bibitem[Song et~al.(2020)Song, Meng, and Ermon]{song2020denoising}
Jiaming Song, Chenlin Meng, and Stefano Ermon.
\newblock Denoising diffusion implicit models.
\newblock \emph{arXiv preprint arXiv:2010.02502}, 2020.

\bibitem[Teed and Deng(2020)]{teed2020raft}
Zachary Teed and Jia Deng.
\newblock Raft: Recurrent all-pairs field transforms for optical flow.
\newblock In \emph{ECCV}, pages 402--419. Springer, 2020.

\bibitem[Wan et~al.(2025)Wan, Wang, Ai, Wen, Mao, Xie, Chen, Yu, Zhao, Yang, et~al.]{wan2025wan}
Team Wan, Ang Wang, Baole Ai, Bin Wen, Chaojie Mao, Chen-Wei Xie, Di Chen, Feiwu Yu, Haiming Zhao, Jianxiao Yang, et~al.
\newblock Wan: Open and advanced large-scale video generative models.
\newblock \emph{arXiv preprint arXiv:2503.20314}, 2025.

\bibitem[Wu and De~la Torre(2023)]{wu2023latent}
Chen~Henry Wu and Fernando De~la Torre.
\newblock A latent space of stochastic diffusion models for zero-shot image editing and guidance.
\newblock In \emph{ICCV}, pages 7378--7387, 2023.

\bibitem[Xing et~al.(2025)Xing, Groh, Karaoglu, Gevers, and Bhattad]{xing2025luminet}
Xiaoyan Xing, Konrad Groh, Sezer Karaoglu, Theo Gevers, and Anand Bhattad.
\newblock Luminet: Latent intrinsics meets diffusion models for indoor scene relighting.
\newblock In \emph{CVPR}, pages 442--452, 2025.

\bibitem[Yang et~al.(2025)Yang, Zhou, Zhao, Tao, and Loy]{yang2025matanyone}
Peiqing Yang, Shangchen Zhou, Jixin Zhao, Qingyi Tao, and Chen~Change Loy.
\newblock Matanyone: Stable video matting with consistent memory propagation.
\newblock In \emph{CVPR}, pages 7299--7308, 2025.

\bibitem[Zeng et~al.(2025)Zeng, Liu, Feng, Miao, Gao, Qu, Zhang, Wang, and Yuan]{zeng2025lumen}
Jianshu Zeng, Yuxuan Liu, Yutong Feng, Chenxuan Miao, Zixiang Gao, Jiwang Qu, Jianzhang Zhang, Bin Wang, and Kun Yuan.
\newblock Lumen: Consistent video relighting and harmonious background replacement with video generative models.
\newblock \emph{arXiv preprint arXiv:2508.12945}, 2025.

\bibitem[Zhang et~al.(2021)Zhang, Zhang, Wu, Yu, and Xu]{zhang2021neural}
Longwen Zhang, Qixuan Zhang, Minye Wu, Jingyi Yu, and Lan Xu.
\newblock Neural video portrait relighting in real-time via consistency modeling.
\newblock In \emph{ICCV}, pages 802--812, 2021.

\bibitem[Zhang et~al.(2023)Zhang, Rao, and Agrawala]{zhang2023adding}
Lvmin Zhang, Anyi Rao, and Maneesh Agrawala.
\newblock Adding conditional control to text-to-image diffusion models.
\newblock In \emph{ICCV}, pages 3836--3847, 2023.

\bibitem[Zhang et~al.(2025)Zhang, Rao, and Agrawala]{zhang2025scaling}
Lvmin Zhang, Anyi Rao, and Maneesh Agrawala.
\newblock Scaling in-the-wild training for diffusion-based illumination harmonization and editing by imposing consistent light transport.
\newblock In \emph{ICLR}, pages 1--12, 2025.

\bibitem[Zheng et~al.(2024{\natexlab{a}})Zheng, Gao, Fan, Liu, Laaksonen, Ouyang, and Sebe]{BiRefNet}
Peng Zheng, Dehong Gao, Deng-Ping Fan, Li Liu, Jorma Laaksonen, Wanli Ouyang, and Nicu Sebe.
\newblock Bilateral reference for high-resolution dichotomous image segmentation.
\newblock \emph{CAAI Artificial Intelligence Research}, 2024{\natexlab{a}}.

\bibitem[Zheng et~al.(2024{\natexlab{b}})Zheng, Gao, Fan, Liu, Laaksonen, Ouyang, and Sebe]{zheng2024bilateral}
Peng Zheng, Dehong Gao, Deng-Ping Fan, Li Liu, Jorma Laaksonen, Wanli Ouyang, and Nicu Sebe.
\newblock Bilateral reference for high-resolution dichotomous image segmentation.
\newblock \emph{arXiv preprint arXiv:2401.03407}, 2024{\natexlab{b}}.

\bibitem[Zhou et~al.(2025)Zhou, Bu, Ling, Zhang, Wu, Huang, Li, Dong, Zang, Cao, et~al.]{zhou2025light}
Yujie Zhou, Jiazi Bu, Pengyang Ling, Pan Zhang, Tong Wu, Qidong Huang, Jinsong Li, Xiaoyi Dong, Yuhang Zang, Yuhang Cao, et~al.
\newblock Light-a-video: Training-free video relighting via progressive light fusion.
\newblock \emph{arXiv preprint arXiv:2502.08590}, 2025.

\end{thebibliography}
}

\clearpage
\setcounter{page}{1}
\maketitlesupplementary

\appendix
\renewcommand{\thesection}{\Alph{section}}
\renewcommand{\thefigure}{\Alph{figure}}
\renewcommand{\thetable}{\Alph{table}}
\begin{figure*}[h]
    \centering
    \includegraphics[width=0.97\textwidth, page=2]{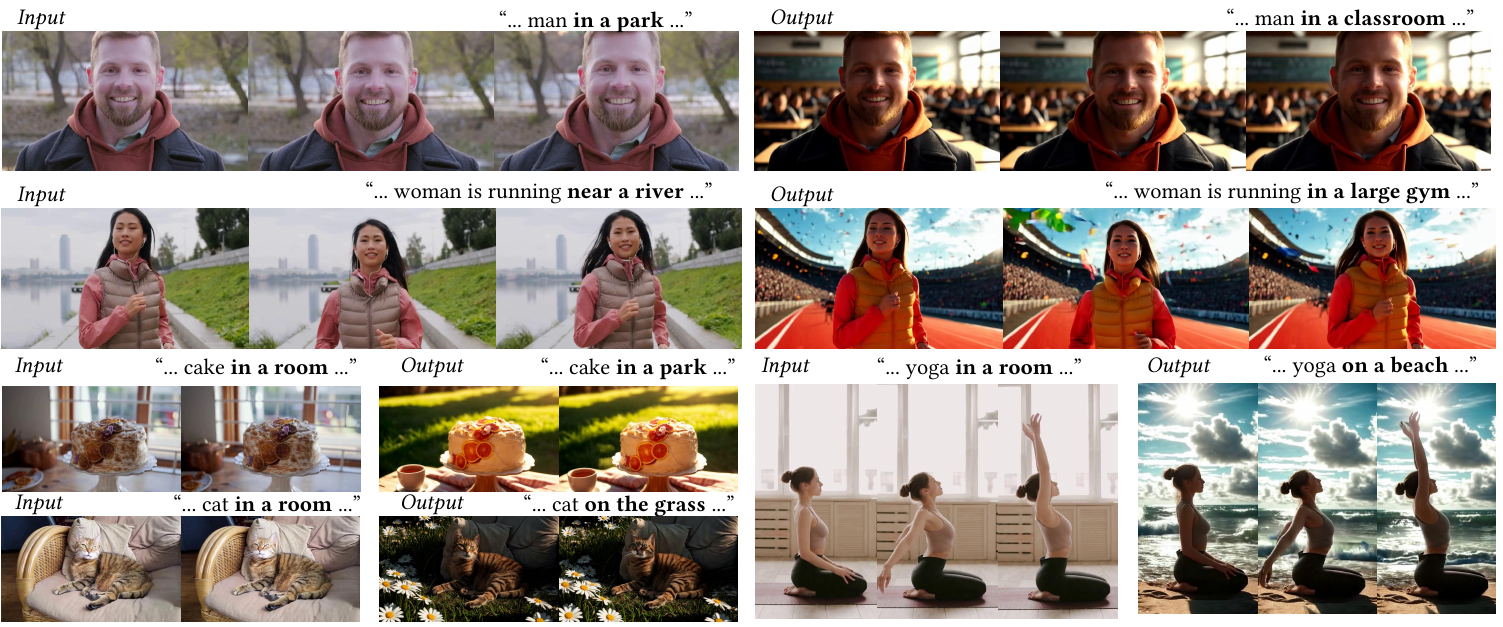}
    \caption{Additional Results of FlowPortal. Refer to our project page for video demonstrations.}
    \label{fig:supmore}
\end{figure*}

\begin{figure*}[h]
    \centering
    \includegraphics[width=0.92\textwidth]{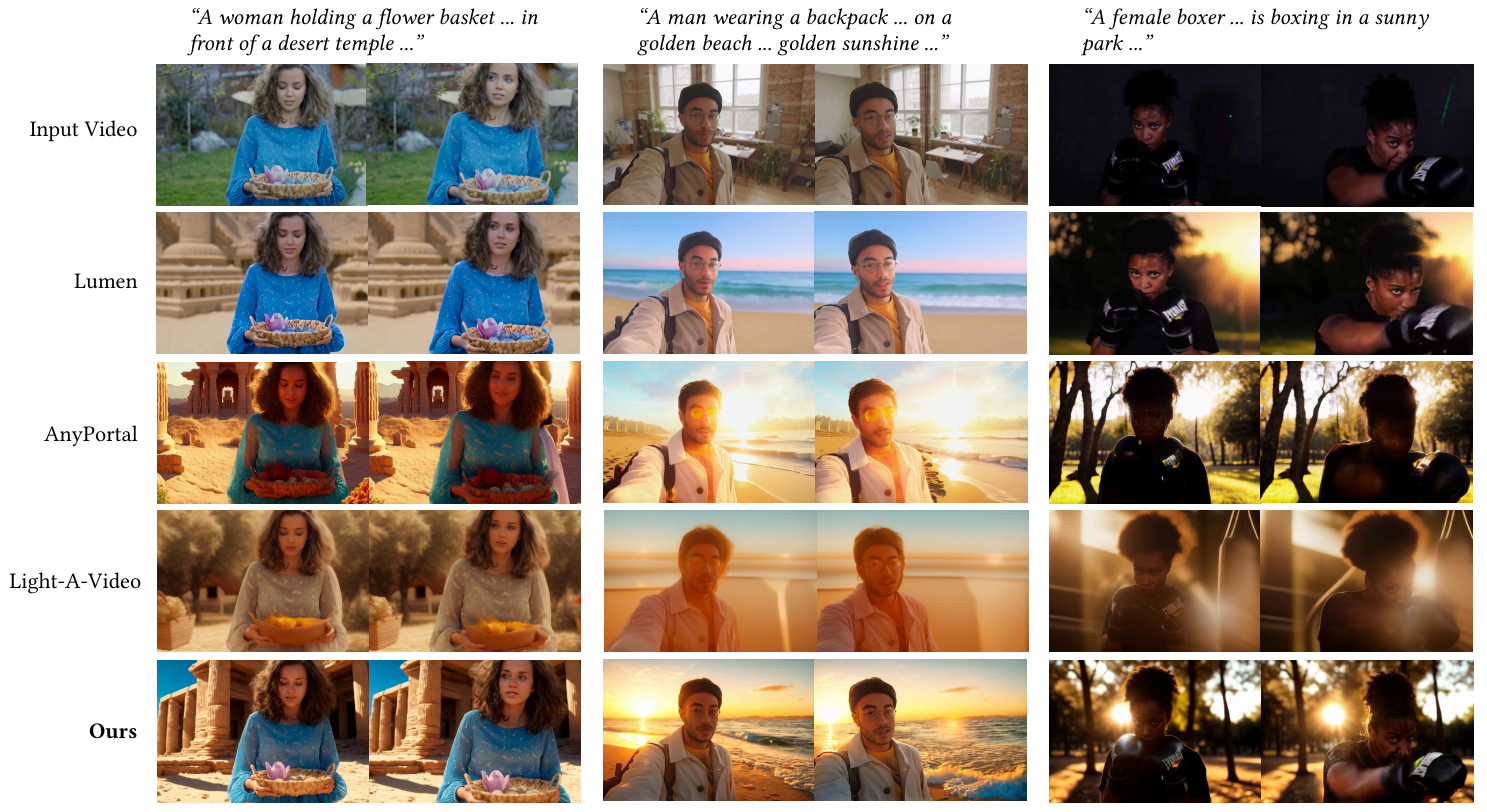}
    \caption{Additional Comparison Results. Refer to our project page for video demonstrations.}
    \label{fig:supcomp}
\end{figure*}

\section*{Appendix}

This supplementary material provides additional qualitative results, methodological details, extended analyses, and user study descriptions. Specifically, we present extra visual results and comparisons in Sec.~\ref{sec:a}, derive Residual-Corrected Flow from FlowEdit in Sec.~\ref{sec:b}, and describe the full algorithmic pipeline in Sec.~\ref{sec:c}. Extensions of our framework, including full-scene relighting and user-provided background control, are provided in Sec.~\ref{sec:d} and Sec.~\ref{sec:e}. We discuss limitations and failure cases in Sec.~\ref{sec:f}, and the user study is demonstrated in Sec.~\ref{sec:g}.

\section{Addtional Results}
\label{sec:a}
We present additional visual results and comparison results in Fig.~\ref{fig:supmore} and Fig.~\ref{fig:supcomp}, respectively.~For video demonstrations, please refer to our project page: https://gaowenshuo.github.io/FlowPortalProject/.

\section{From FlowEdit to Residual-Corrected Flow}
\label{sec:b}

As shown in Sec.~3.6 of our main paper, our method is partially inspired by FlowEdit~\cite{kulikov2025flowedit}. Specifically, assuming FlowEdit use the same global noise \(\epsilon\) at every timestep:  
\begin{align}
\label{eq:flowedit}
V^{\text{FE-edit}}_{t}(z^{\text{FE-edit}}_{t}) &= V^{\text{tar}}_{t}(z^{\text{FE-pred}}_{t}) - V^{\text{src}}_{t}(z_{t}),\\
z_{t} &= (1 - t)z_0 + t\epsilon,\\
z^{\text{FE-pred}}_{t} &= z_{t} + z^{\text{FE-edit}}_{t} - z_0,
\end{align}
where \(t \in [0,1]\) and \(z^{\text{FE-edit}}_{1} = z_0\). Then we can derive:  
\begin{align}
z^{\text{FE-pred}}_{1} &= \epsilon, \\
V^{\text{FE-pred}}_{t}(z^{\text{FE-pred}}_{t}) &= V_t(z_t) + V^{\text{FE-edit}}_{t}(z^{\text{FE-edit}}_{t}) - 0 \\
&= V_t(z_t) + V^{\text{tar}}_{t}(z^{\text{FE-pred}}_{t}) - V^{\text{src}}_{t}(z_{t}). \nonumber
\end{align}

Note that \(V_t(z_t)\) is a constant velocity:  
\begin{align}
V_t(z_t) = \frac{z_0 - \epsilon}{1 - 0} = V_0, 
\end{align}
so that  
\begin{align}
V^{\text{FE-pred}}_{t}(z^{\text{FE-pred}}_{t}) = V^{\text{tar}}_{t}(z^{\text{FE-pred}}_{t}) + (V_0 - V^{\text{src}}_{t}(z_{t})),
\end{align}
which shows that \(z^{\text{FE-pred}}_{t}\) can be interpreted as a generation process starting from \(\epsilon\) and evolving along \(V^{\text{FE-pred}}_{t}\). We then rename this variable as \(z^{\text{edit}}_{t}\) and track its trajectory. Furthermore, we observe that \(V_0 - V^{\text{src}}_{t}(z_{t})\) can be interpreted as a residual velocity corresponding to reconstructing \(z_0\) using the source prompt, which we denote as:
\begin{align}
V^{\text{res}}_{t}(z_t) = V_0 - V^{\text{src}}_{t}(z_{t}),
\end{align}
Finally, we obtain:  
\begin{align}
V^{\text{edit}}_{t}(z^{\text{edit}}_{t}) = V^{\text{tar}}_{t}(z^{\text{edit}}_{t}) + V^{\text{res}}_{t}(z_{t}),
\end{align}
which corresponds to our method.

Essentially, $V^{\text{res}}_{t}$ is a residual velocity that maintains the structure and content of the original video, ensuring that the editing process preserves structural consistency.
Its value is equivalent to the finite difference with respect to \(t\) of the difference between the original video \(z_0\) and the result of a single-step denoising applied to \(z_0\) after adding noise at level \(t\).
Therefore, \(V^{\text{res}}_t\) actually carries the precise structural information of the original video, making it particularly suitable for video relighting tasks.

Vanilla FlowEdit actually selects \(n\) different noise samples at each step of the generation process and averages the resulting outputs, which leads to stability but 
blurring in the generated results. To produce a clear background, we discard this multi-noise sampling procedure.

This reinterpretation offers the following advantages over the original FlowEdit: First, our reinterpretation allows the background to remain a purely generative process. By simply multiplying \(V^{\text{res}}_t\) with a foreground mask \(M\), the background is prevented from being influenced by the structural information of the original video, thereby avoiding artifacts, as shown in Fig.~8  in our main paper. Second, to reduce additional computational overhead, our reinterpretation enables the structural residual information in \(V^{\text{res}}_t\) to be reused every \(r\) steps, reducing the total number of steps from \(2T\) to \((1 + 1/r)T\) without significantly degrading generation quality, as also illustrated in Fig.~8 in our main paper.

\section{Full Algorithm}
\label{sec:c}

In this section, we present the detailed implementation of our proposed framework.

\subsection{Decoupled Condition Design}

We first construct the condition inputs, including text prompts, reference frames, and shared structural information.  
To obtain the foreground mask of the input video, denoted as $M$, we apply BiRefNet~\cite{zheng2024bilateral} on the first frame and propagate it across the sequence using MatAnyone~\cite{yang2025matanyone}.  
Given the mask, the reference frames and structural conditions are prepared as follows:

\begin{algorithm*}[t]
\caption{Residual-Corrected Flow for Video Relighting}
\label{alg:rcf}
\begin{algorithmic}[1]
\State \textbf{Input:} Initial noise $\epsilon$, input video $z_0$, timestep schedule $0 = t_0 < t_1 < \dots < t_N = 1$, velocity field $V^{\text{src}}_t$ and $V^{\text{tar}}_t$, reusing steps $r$, \textit{High-Frequency Transfer} intensity $\lambda$, foreground mask $M$
\State \textbf{Output:} Relit video $z^{\text{edit}}_0$
\vspace{2mm}

\State $V_0 \gets \epsilon - z_0$
\State $z^{\text{edit}}_{t_N} \gets \epsilon$

\For{$i = N, N-1, \dots, 1$}
    \State $z_{t_i} \gets (1 - t_i) z + t_i \epsilon$
    \State $z_{t_{i-1}} \gets (1 - t_{i-1}) z + t_{i-1} \epsilon$
    \If{$(N - i) \bmod r == 0$} \Comment{calculate new residual for reusing in next $r$ steps}
        \State $V^{\text{res}}_{t_i}(z_{t_i}) \gets V_0 - V^{\text{src}}_{t_i}(z_{t_i})$
        \State $t_{\text{last}} \gets t_i$
    \EndIf
    \State $V^{\text{edit}}_{t_i}(z^{\text{edit}}_{t_i}) \gets M \cdot V^{\text{res}}_{t_{\text{last}}}(z_{t_i}) + V^{\text{tar}}_{t_i}(z^{\text{edit}}_{t_i})$
    \State $z^{\text{edit}}_{t_{i-1}} \gets z^{\text{edit}}_{t_i} + (t_i - t_{i-1}) V^{\text{edit}}_{t_i}(z^{\text{edit}}_{t_i})$
    \State $z^{\text{edit}}_{t_{i-1}} \gets 
        \mathrm{LF}(z^{\text{edit}}_{t_{i-1}}) 
        + \lambda M \cdot \mathrm{HF}(z_{t_{i-1}}) 
        + (1 - \lambda M) \cdot \mathrm{HF}(z^{\text{edit}}_{t_{i-1}})$
\EndFor
\end{algorithmic}
\end{algorithm*}

\begin{itemize}
    \item \textbf{Reference frame for the source condition.}  
    We directly use the first frame of the input video as the reference.  
    Note that, in Wan2.1~\cite{videoxfun}, where the first frame is not strictly required, any frame can serve as the reference.
    
    \item \textbf{Reference frame for the target condition.}  
    We employ IC-Light~\cite{zhang2025scaling} to relight the selected source frame according to the target textual description.  
    The inputs to IC-Light include the chosen video frame, its corresponding foreground mask, and the target prompt.
    
    \item \textbf{Shared structural information.}  
    We extract Canny edges ($C$), HED boundaries ($H$), and depth maps ($D$) from the original video~\cite{zhang2023adding}, and empirically combine them as $(0.25C + 0.25H + 0.5D) \cdot M$.  
    The edge-based components (Canny, HED) preserve fine structural details, while the depth map encourage structural consistency without over-constraining the model.  
    This structural condition is shared between both source and target conditions.
\end{itemize}

\subsection{Inference Algorithm}

Starting from a global Gaussian noise $\epsilon$, we perform denoising using Wan2.1~\cite{wan2025wan, videoxfun} integrated with our proposed method.  
At each timestep $t$, the model predicts the velocity fields $V^{\text{src}}_t$ and $V^{\text{tar}}_t$ under the source and target conditions, respectively.  
The complete inference procedure is shown in Algorithm~\ref{alg:rcf}.

\section{Extension: Full Scene Relighting}
\label{sec:d}

\begin{figure}[h]
    \centering
    \includegraphics[width=0.97\linewidth]{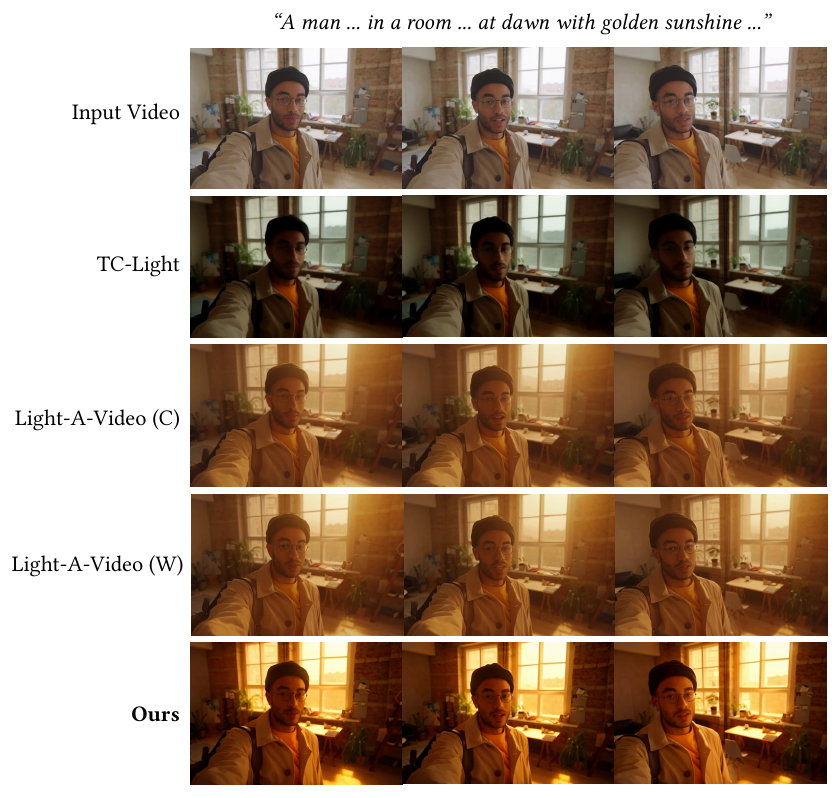}
    \caption{\textbf{Full Scene Relighting comparison.} Our method is capable of relighting entire scene while preserving background structural coherence. Compared with TC-Light and Light-A-Video with different backbones, our approach produces the most natural illumination and preserves scene structure most faithfully.}
    \label{fig:ext1}
\end{figure}

In our method, removing the masking mechanism in condition preparation, Residual-Corrected Flow, and High-Frequency Transfer enables full scene relighting, allowing the model to adjust illumination across the entire scene while preserving background structure. We provide a visual comparison of full scene relighting against TC-Light~\cite{liu2025tc} and Light-A-Video~\cite{zhou2025light} with different backbones, in Fig.~\ref{fig:ext1}.
Notably, our approach achieves the most natural illumination while maintaining the highest level of structural consistency across the entire scene.

\section{Extension: User-Provided background}
\label{sec:e}

\begin{figure}[h]
    \centering
    \includegraphics[width=0.97\linewidth]{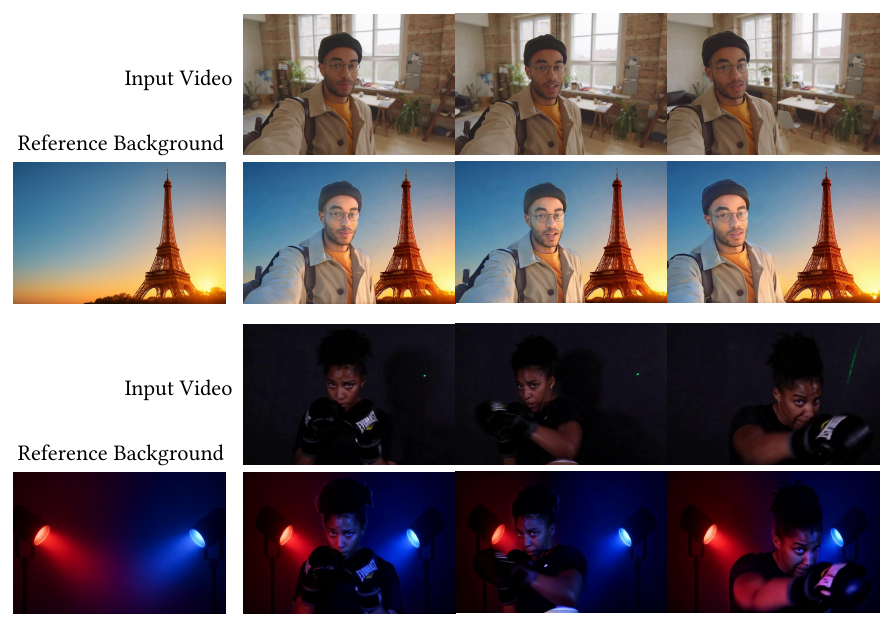}
    \caption{\textbf{Scene-specified relighting.} Our method supports user-specified background scenes by feeding the target scene image into IC-Light during reference-frame preparation, enabling foreground relighting consistent with the desired scene.}
    \label{fig:ext2}
\end{figure}

In some user scenarios, the user may wish to specify the background scene of the generated video, and our method naturally supports this type of task. During condition preparation, we input the user-specified scene image into IC-Light when generating the reference frame, enabling our method to produce a relit video whose foreground matches the lighting of the specified scene as iluustrated in  Fig.~\ref{fig:ext2}.

\section{Limitation}
\label{sec:f}

\begin{figure}[h]
    \centering
    \includegraphics[width=0.97\linewidth]{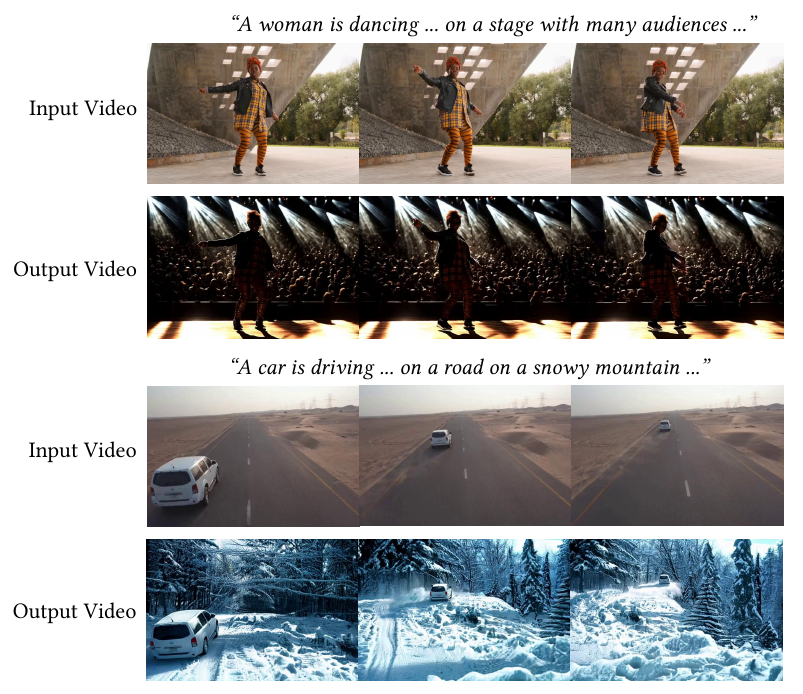}
    \caption{\textbf{Failure Cases}. Although our method is capable of handling a wide range of video relighting and background replacement scenarios, it can still be constrained by the generation capacity of its two base models in extremely complex situations.
    In the first example, the IC-Light model fails to generate a complex backlit stage scene containing many audience, .
    In the second example, IC-Light produces a reference frame whose scene layout does not align with the actual driving trajectory of the vehicle. This mismatch restricting the video model’s ability to produce a reasonable and structurally coherent output.}
    \label{fig:limit}
\end{figure}

Although our method can handle a wide range of video relighting and background replacement scenarios, it may still be limited in extremely complex cases by the generation capabilities of the two base models, as shown in Fig.~\ref{fig:limit}.

\section{User Study Details}
\label{sec:g}
\begin{figure*}[htbp]
    \centering
    \includegraphics[width=0.97\textwidth]{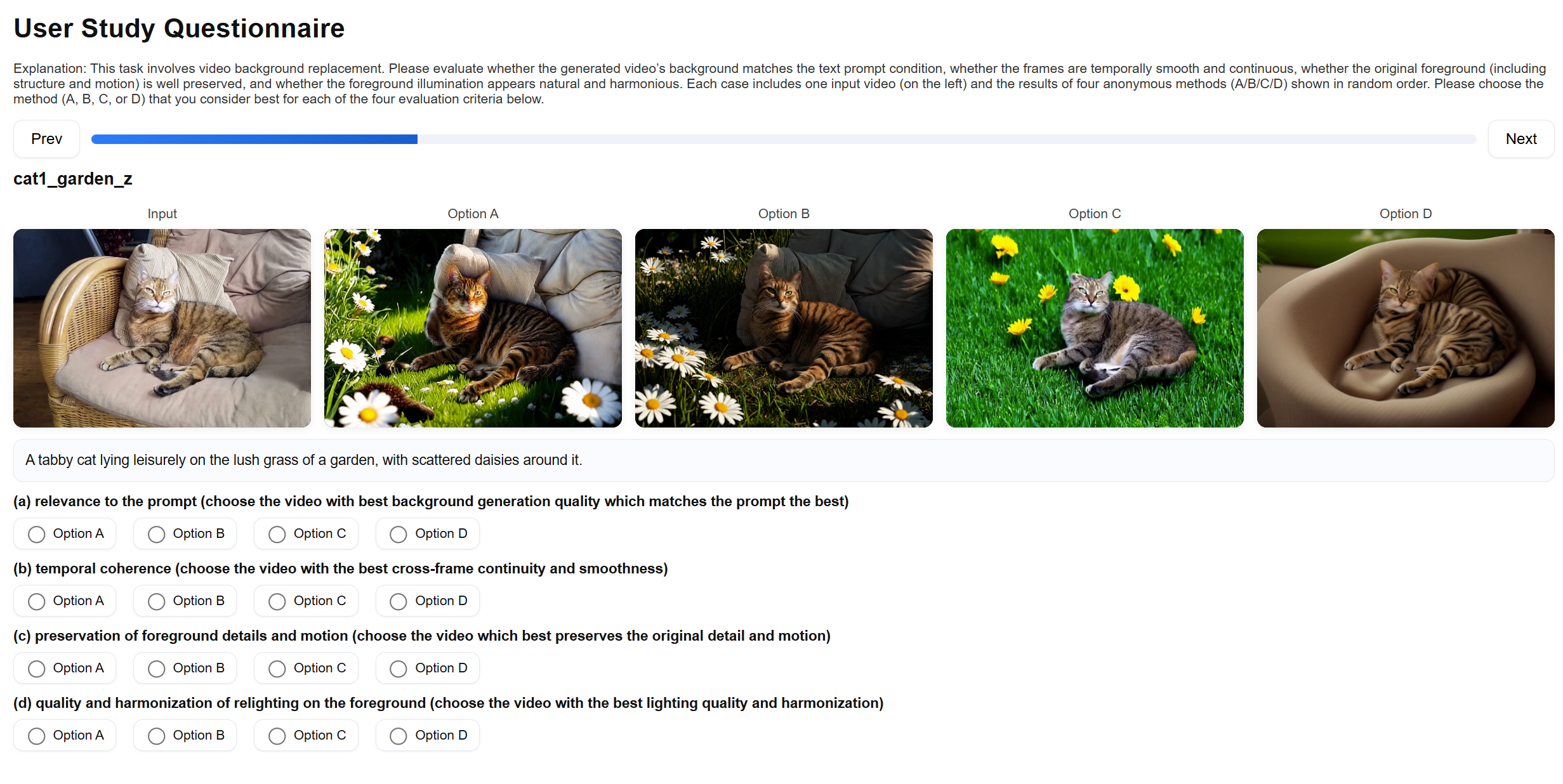}
    \caption{A screenshot of the User Study interface.}
    \label{fig:user}
\end{figure*}

The user study in our main paper is conducted with 24 invited participants, who were asked to complete a questionnaire comparing our method with AnyPortal~\cite{gao2025anyportal}, Light-A-Video~\cite{zhou2025light}, and Lumen~\cite{zeng2025lumen} on a dataset of randomly-selected 17 video–prompt pairs. A screenshot of the evaluation interface is shown in Fig.~\ref{fig:user}.

\end{document}